# A Structured Review of Literature on Uncertainty in Machine Learning & Deep Learning

Fahimeh Fakour     Ali Mosleh     Ramin Ramezani

## Abstract

The adaptation and use of Machine Learning (ML) in our daily lives has led to concerns in lack of transparency, privacy, reliability, among others. As a result, we are seeing research in niche areas such as interpretability, causality, bias and fairness, and reliability. In this survey paper, we focus on a critical concern for adaptation of ML in risk-sensitive applications, namely understanding and quantifying uncertainty. Our paper approaches this topic in a structured way, providing a review of the literature in the various facets that uncertainty is enveloped in the ML process. We begin by defining uncertainty and its categories (e.g., aleatoric and epistemic), understanding sources of uncertainty (e.g., data and model), and how uncertainty can be assessed in terms of uncertainty quantification techniques (Ensembles, Bayesian Neural Networks, etc.). As part of our assessment and understanding of uncertainty in the ML realm, we cover metrics for uncertainty quantification for a single sample, dataset, and metrics for accuracy of the uncertainty estimation itself. This is followed by discussions on calibration (model and uncertainty), and decision making under uncertainty. Thus, we provide a more complete treatment of uncertainty: from the sources of uncertainty to the decision-making process. We have focused the review of uncertainty quantification methods on Deep Learning (DL), while providing the necessary background for uncertainty discussion within ML in general. Key contributions in this review are broadening the scope of uncertainty discussion, as well as an updated review of uncertainty quantification methods in DL.

## Introduction: Machine Learning in Our Lives Today

In our everyday lives, we continue to see the growing role of artificial intelligence (AI) and ML. ChatGPT can answer a question, provide clarifying examples, and even help plan your vacation. Apple's Siri, Amazon's Alexa, and Google's Nest, all attest to the extent of AI and ML influence on our daily lives and comfort. The fact that your navigation system makes your commute a little easier, or the recommendation of music matches (or elevates) your mood, are only a few of the myriad ways by which ML and AI are influencing our lives, for better (completing tasks more efficiently) or worse (e.g., un-useful recommendations).

Despite the impressive usage of ML in various fields, it is more slowly adapted in risk and safety-critical settings [1]. While it may not be critical to have a recommender system that provides you with just the right music track, there are many applications in which the prediction results of a ML algorithm can have critical implications, such as autonomous vehicles, drug development, and medical diagnosis.

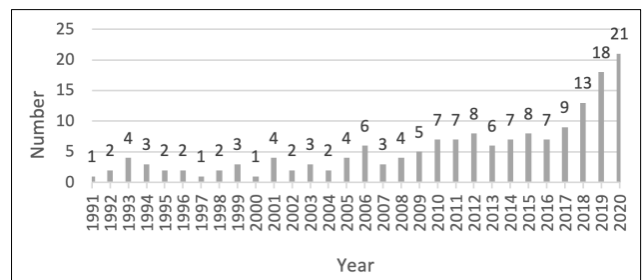

*Figure 1) (borrowed from [2]) Number of papers published on topic of uncertainty handling uncertainty in medical data (1991-2020)*

**Literature Landscape**

In this review, uncertainty discussion and quantification methods are involved in a variety of ML applications: weather forecasting, IoT, NLP, computer vision, medicine, drug development, etc. In addition to these applications, there are also algorithms and methodologies in ML that make use of uncertainty handling and estimation (such as active learning, reinforcement learning, out-of-distribution detection, etc.). There is increasing interest in uncertainty in ML, which is reflected in the growth of literature. Figure 1 illustrates this increase in research in the medical data field [2]. The dynamic and fast pace of research development in uncertainty in ML is not limited to the medical field, as there is emerging literature in various other areas, such as usage of uncertainty estimation techniques for LLMs [3][4].

This paper offers a review of the literature pertaining to uncertainty in machine learning (ML), with a particular focus on uncertainty quantification (UQ) techniques for deep learning (DL). It represents a necessary step towards organizing and synthesizing various discussions surrounding uncertainty. We advocate for a holistic approach for discussing uncertainty within the ML context, which aims to provide a clear and comprehensive grasp of its multifaceted implications for both practitioners and researchers. We highlight several key aspects that are crucial in addressing uncertainty in ML: 1) distinguishing between the categories of uncertainty (aleatoric vs epistemic), 2) identifying sources of uncertainty (data and model), 3) reviewing metrics for uncertainty quantification and evaluating UQ accuracy, 4) exploring UQ techniques, and 5) examining decision-making processes under uncertainty. To the best of our knowledge, this paper is the first attempt to systematically organize and address uncertainty discussions in this manner. Furthermore, we present an updated review of UQ techniques in DL and provide a high-level overview of how the various techniques differ. We have refrained from a review of the wealth of application literature on UQ, as other papers (such as [5][6]) provide comprehensive coverage in this area.

The importance of addressing uncertainty has been acknowledged in the usage of ML, especially for risk-sensitive settings. We argue that a holistic uncertainty approach aids in identifying some of the ambiguities for practitioners, allowing for comprehensive treatment of uncertainty. For instance, a look at the different uncertainty categories and sources in their application and data, may aid the practitioner in choosing a proper UQ technique, thus allowing them to identify areas of high aleatoric/epistemic uncertainty. To this end, there is a growing amount of literature in quantifying the aleatoric and epistemic components of the total uncertainty. On a similar note, the choice of metric and addressing the accuracy of the prediction uncertainty gains importance, as well as discussions of decision-making context and considerations of decision-making theories. Putting these different pieces together - in a holistic view at uncertainty- is an important initial step in providing a comprehensive framework for measuring, communicating, and understanding uncertainty in the ML process. Most of the discussions in this paper consider a supervised learning setting, meaning that the data samples have a true value, and the model output can be evaluated through the consideration of the true value.

The paper is structured as follows: First, we provide the motivation for addressing and assessment of uncertainty. Having thus established the importance of this subject, especially in risk-sensitive settings, this literature review proceeds to approach uncertainty in ML in a holistic manner: looking at the definition and categorization of uncertainty, the sources of uncertainty, establishing metrics for evaluating uncertainty in ML, providing an updated review of UQ techniques, and discussion of decision making under uncertainty. We conclude with a discussion of future work.

## 1. Motivational background

It can be argued that the vast adaptation of ML is largely due to its ability to provide useful predictions. Understandably, part of the ML process involves measuring the performance of the model, using that as an indication of the model's ability to provide adequate predictions for the task at hand. Depending on the type of data, researchers may choose to evaluate the performance of ML models with such measures as error rate (e.g., Mean Absolute Error (MAE), Mean Squared Error (MSE) for regression tasks; logarithmic loss and Area Under the Curve (AUC) for classification tasks [7]). It is important to note that the choice of performance metric(s) can vary based on the data (e.g., balanced/imbalanced property of the dataset). As part of addressing uncertainty, ongoing research seeks to provide a metric and means of assessing uncertainty in a quantitative manner. However, many standard ML practices and metrics do not provide substantial and/or accurate insight into the uncertainty on the predicted output of the model [8].

In this section, we outline some of the shortcomings of using common, standard ML metrics for uncertainty quantification/estimation in standard ML practice. While some of these metrics (such as standard error) may sufficiently communicate uncertainty for some applications, there is a requirement for more precise and fine-grained estimations of uncertainty particularly for risk-sensitive models.

### 1.1 Why Standard Error (SE) is not Enough

While some practitioners provide an interval of $(y_{prediction} \pm SE)$ as a measurement of the confidence/uncertainty of the prediction, this approach considers the standard error a depiction of the average uncertainty (interpreting mistakes as uncertainty points). For risk-sensitive models, there is a requirement for a more fine-grained uncertainty estimation: uncertainty needs to be estimated for each input sample. Standard error fails to satisfy this requirement [8].

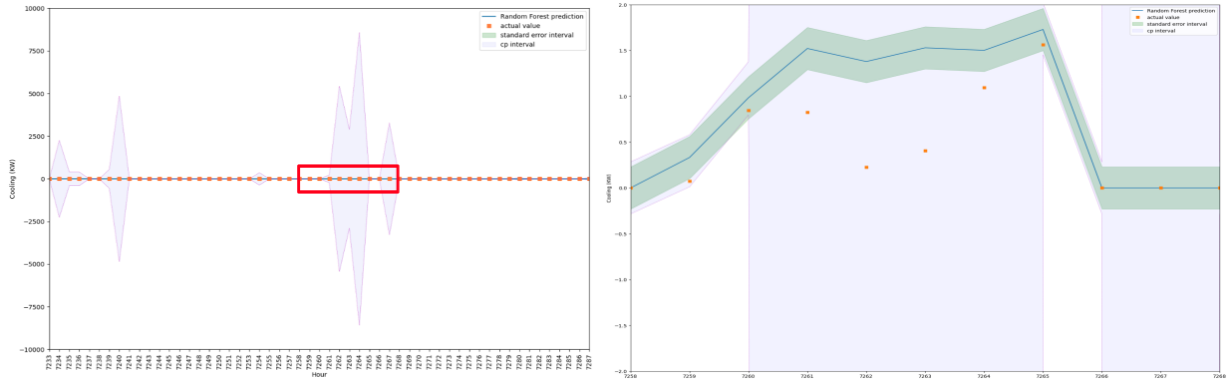

*Figure 2) Example illustrating the difference in conformal prediction intervals (purple shaded areas) and standard error (green shaded). (Figure left:) shows the large conformal prediction intervals in highlighted area (red box). This hints at high uncertainty for this prediction area. (Figure right:) close-up of highlighted section, shows that the actual values (orange points) in the area with larger CP intervals indeed fall further from prediction line (blue line). It is noteworthy that the actual values in this area fall outside the standard error interval.*

Figure 2 provides an example of the prediction intervals with uncertainty quantification in interval predictions (conformal prediction, in this example) and standard error. The green shaded area depicts the standard error interval surrounding the ML predictions. As illustrated in the figure, Conformal Prediction (CP), in contrast with SE (which is constant for all the samples), provides a more detailed view of the uncertainty by providing a prediction interval *per sample*. These intervals are tailored for each prediction, with the very large intervals communicating the high uncertainty associated with the prediction. Standard error, unfortunately, lacks this added insight. This figure illustrates the added value of conformal prediction: we see that in some of the areas in which the actual value (ground truth) falls outside the standard error (green area), the conformal prediction interval is very large, pointing towards high uncertainty in the prediction. In such cases, equipped with the added knowledge provided by the large interval from CP, researchers are made aware of the high uncertainty, and may avoid making an erroneous decision.

## 1.2 Why Softmax is not Enough

Softmax is a mathematical function that can be used in the output layer of a neural network for multi-class classification. This function takes the raw scores received at the output layer and normalizes these scores into a probability distribution over the different classes. It has been ascertained that high probabilities do not necessarily provide the best results; this is exemplified in the hallucination problem of Large Language Models (LLMs). Recent literature investigates hallucination in LLMs using uncertainty estimation techniques [3][4]. In this section, we provide three examples why softmax outputs are not an adequate measure of the model's uncertainty/confidence: 1) stochastic nature of training models; 2) softmax with combined labels; 3) softmax with out-of-distribution (OOD) samples.

## Stochasticity in Model Training

When there is randomness in model training and there is a complex cost function, the learned model parameters may vary (change in local minima). Considering the stochastic nature of training such models, running the same model design with different parameter initialization may result in different models, which in turn, yield different softmax probabilities for the same sample. This is easily seen in the use of ensembles. Consider the following MNIST classification example: a sample gives a prediction probability (softmax output) of [0.1, 0.05, 0.3, 0.18, 0.05, 0.23, 0.05, 0.02, 0.02]. We may be misled to think that the model is not very confident about this classification (max probability is only 0.3 for class label 2); however, this is not enough information to ascertain the predictive uncertainty. In contrast, consider that this probability scoring example is part of an ensemble of 10 models, and –for this sample– all the ensemble outputs give a probability of 0.3 for label 2. In this case, we can make the claim that the model actually has low uncertainty (is very confident) that the probability of this sample being in class label 2 is 0.3. This example illustrates that we cannot rely on a single softmax score for an accurate understanding of the predictive uncertainty.

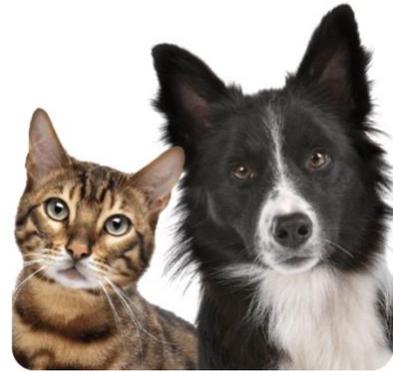

*Figure 3) Image of Cat and Dog, an example of combined labels. When providing this sample to an ML model trained on images of cats/dogs, the model would find it difficult to communicate the identification of cat and dog using softmax possibilities.*

## Combined labels

We provide a toy-example to highlight the limitation of the softmax probabilities as a measure of confidence[1]. Note that the softmax probabilities must sum to 1, so it is "forced" to choose from the given classes and doesn't consider alternative outcomes. Consider an image classification system that is trained for images of cats and dogs. Now the input sample to this trained model contains both a cat and a dog (Figure 3). The image contains both objects that the model is trained to classify; thus, it identifies the features of cat, as well as dog, in the image. The model gives softmax probability scores of [0.5, 0.5]. An initial reaction to these probability scores would lead to the belief that the model believes that there is a 50-50 chance that the image contains a cat or a dog. The [0.5, 0.5] prediction probability may lead us (mistakenly) to think that the model is highly uncertain and undecided between the two classes. However, this may not be the case: the model might be very confident it sees the features of a cat and a dog; given the opportunity, it could have attributed high probability for both labels (e.g. [0.9, 0.9]). However, the model is constrained by softmax (thus giving [0.5, 0.5]).

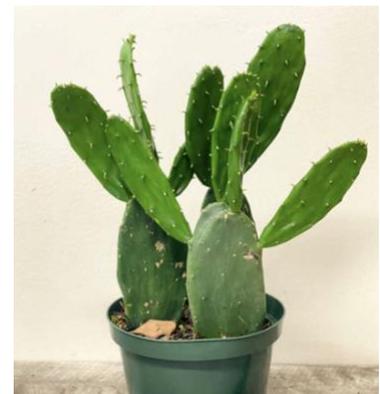

*Figure 4) Image of Cactus, example of OOD sample. For the dog-cat trained model, the model would have to classify the image as either cat or dog.*

---

[1] Images are courtesy of google images.

OOD Samples

Now, consider that the cat-dog trained model is given a sample containing just a cactus (Figure 4); it gives us softmax probability scores of [0.15, 0.85]. In this case, due to the limitations and definition of softmax, the model has no way of communicating that it 'doesn't know' about this sample (i.e. hasn't seen anything similar in its training). Ultimately, [9] points out that- for OOD points- standard softmax NNs extrapolate arbitrarily and suffer from feature collapse. [9] shows empirically that for a dataset with high epistemic and aleatoric uncertainty, softmax entropy has arbitrary values for OOD samples and is unable to identify in-distribution samples from OOD.

Final Notes on Softmax for Uncertainty Estimation

While we have summarized the aforementioned shortcomings and prohibitions of using softmax as a measure of confidence/uncertainty, many studies investigate softmax as an indication of confidence/uncertainty. In an interesting study, [10] focuses on the strengths and shortcomings of using softmax for confidence estimation, particularly that softmax does 'reasonably well' in providing an uncertainty estimate for non-adversarial OOD detection, in datasets with low aleatoric uncertainty [10]. [10] clarifies that in low-dimensional data and shallow networks, the softmax layer doesn't reliably estimate the uncertainty; however, in the case of high-dimensional data and deep networks, the softmax layer is able to identify the OOD samples and map them to low-confidence values. While their findings further highlight that softmax is unable to provide adversarial support and/or provide confidence measures for data with high aleatoric uncertainty, the study provides support on the usefulness of softmax for applications without high risk/reliability concerns. [10] further summarizes this point that "Softmax confidence remains an imperfect measure of uncertainty, and caution should be applied when used in real-world applications." Their study provides mathematical background and possible insights to ways that softmax can assist in OOD detection.

## 2. Uncertainty Definition

Uncertainty is defined broadly as lack of knowledge, whether one is aware of the difference in knowledge from the idealized state or not [8]. Using this initial definition as a stepping-stone, we attempt to understand how uncertainty, as "lack of knowledge," pertains to ML process, its sources, and literature. It is noteworthy that this is a broad, and perhaps vague definition, making discussions sometimes difficult and confusing. The overlapping semantics in measures of reliability, confidence, uncertainty, probabilities, and so on, makes the investigation of uncertainty a difficult one. We found that the majority of our reviewed literature focused on the quantification of uncertainty portrayed in the outcome of the ML process, commonly referred to as predictive uncertainty (also prediction uncertainty). To the best of our knowledge, the majority of uncertainty quantification literature considers predictive uncertainty equivalent to the total uncertainty. Total uncertainty refers to the combination of epistemic and aleatoric uncertainty.

Figure 7 provides a high-level visual representation of the uncertainty discussions and topics in this paper.

## 3. Uncertainty Categorizations: Epistemic vs Aleatoric Uncertainty

In the context of ML (and beyond), uncertainty is categorized as having two types: epistemic and aleatoric. Epistemic uncertainty is understood as the uncertainty that can be alleviated with more data; thus, it is reducible [7]. Aleatoric uncertainty, on the other hand, is uncertainty that captures the stochastic nature and is part of the data/process, hence it is irreducible [7]. Commonly, samples with high aleatoric uncertainty are often referred to as ambiguous samples. Figure 5 provides a visual representation of how aleatoric and epistemic uncertainty manifests in data with uncertainty. Figure 6 shows that these categorizations of uncertainty are indeed fluid. Depending on the setting and design decisions (i.e. feature space complexity of the data), aleatoric uncertainty may turn into epistemic uncertainty and vice versa. This view is also shared by [11], emphasizing that the concepts of aleatoric and epistemic uncertainty are unambiguous when defined within the framework of a model; they further assert that uncertainty addressed as aleatoric in one model, may be considered epistemic in another model.

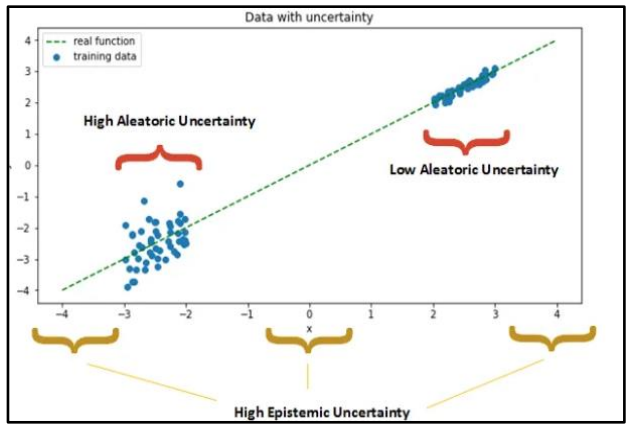 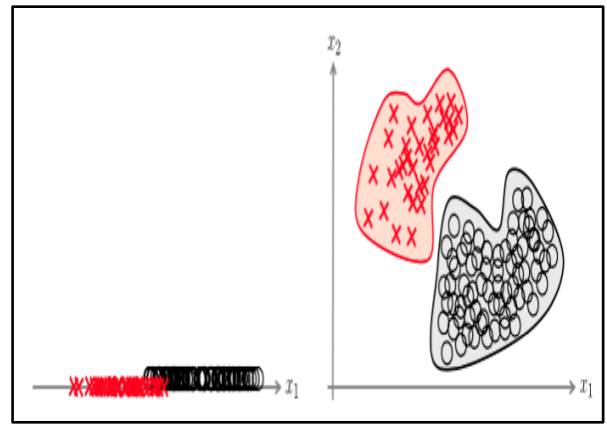

*Figure 5) (borrowed from [12]) Example of high/low aleatoric uncertainty and high/low epistemic uncertainty areas in the data. The areas with high aleatoric uncertainty are ambiguous (overlapping labels). The high epistemic uncertainty areas have little/no data. If a sample falls into this area, the model will make a prediction without having seen similar data.*

*Figure 6) (borrowed from [7]) Changes in the feature space of the data influences the epistemic and aleatoric uncertainties of the data. The plot with only $x_1$ has high aleatoric uncertainty, but low epistemic uncertainty. By encoding data in a higher-dimension space (adding a second feature, $x_2$), the aleatoric uncertainty decreases (classes separate out) but epistemic uncertainty increases (more empty spaces).*

The separation of uncertainty measurements into these two types is a focus of recent literature: attempting to break down and quantify the amount of aleatoric and epistemic uncertainties in the total uncertainty [13]. This is of interest for algorithms and ML methods (such as Active Learning, Reinforcement Learning) and ML applications (such as computer vision, highlighting areas of high epistemic and aleatoric uncertainty in an image) [6][14]. In a recent study, [15] proposes the utilization of aleatoric uncertainty for fairness (addressing bias in predictions).

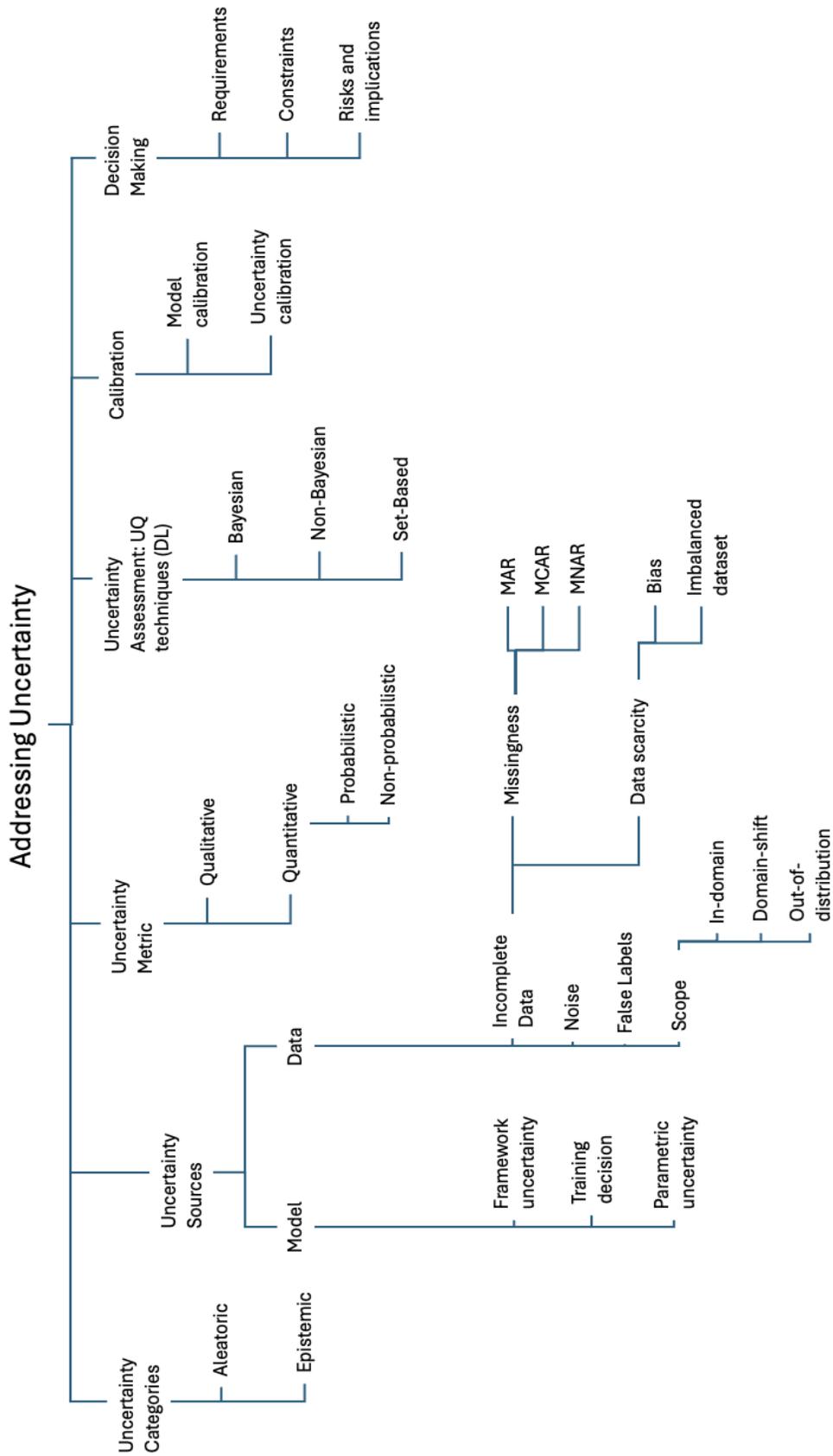

*Figure 7) Uncertainty topics and discussions*

It is noteworthy that some literature consider a third category of uncertainty: distributional. The uncertainty caused by a change from the input data distribution is referred to as distributional uncertainty [16][17]. In other words, distribution uncertainty is the uncertainty caused by distribution shift (explained in Scope section). Distribution shift may be due to covariate/data shift, concept shift, and prior probability shift (label shift). [18] On the other hand, [6] considers distributional uncertainty another source of uncertainty (separate from data and model uncertainty) that can be modeled. It can be argued that OOD detection studies aim to capture the distributional uncertainty [19]. In this paper, however, we have considered distributional uncertainty as part of the data uncertainty (data scope), due to its dependence on the choice of data (namely dataset choice to represent the data distribution intended for the application).

## 4. Uncertainty Sources: Data and Model

Uncertainty sources are the elements that contribute to the total uncertainty in the ML process. When handling uncertainty in ML with a focus on predictive uncertainty, it becomes important to understand the sources of uncertainty, as they may propagate their uncertainty into the predictive uncertainty. The main sources of uncertainty explained here are data uncertainty and model uncertainty.

For an overview of possible sources of uncertainty (as it pertains to the ML process), see the content under uncertainty sources in Figure 7.

### 4.1. Data Uncertainty

ML relies heavily on data: using training data to learn the model, validation data to validate the model accuracy, and test data to evaluate the model. The uncertainty in the data is referred to as data uncertainty. Data uncertainty can have a negative impact on the ML model's performance, reducing its accuracy and reliability. As a result, understanding and managing data uncertainty is important. As early as 1994, [20] talks about the limitations of machine learning that are imposed by the quality of the data, thus highlighting the need for an understanding (and handling) of data uncertainties.

In the realm of Big Data, data quality and uncertainty (referred to as data veracity) is also an area of interesting research [21]. While veracity has challenges specific to a big data setting, there are overlapping data uncertainty concerns such as data variety, high speed of emerging data, and different sources of data [21][22][23][24]. Ultimately, there is a concern that higher amounts of data uncertainty can impact the quality of the data and potentially propagate to higher predictive uncertainty. There are various studies focused on evaluating the quality of data. [25] provides an overview of a data validation system and the challenges encountered when dealing with various data sources. It explains the importance of finding "data errors" early in the ML process, as using erroneous data in the ML pipeline can provide misleading models and predictions.

The data uncertainty and quality of the data are intertwined concepts: higher amounts of data uncertainty can impact the quality of the data and potentially propagate to higher predictive uncertainty. The discussion and remediation of data uncertainties date far beyond current uncertainty quantification discussions. Many of the data uncertainty sources explained in this section are niche fields of study, dealt with extensively in the literature (such as missing data, noise, etc.). These studies have provided the ML community with a wealth of approaches and methods to deal and/or remediate the impact of particular data uncertainties, while focusing mainly on their impact and improvement of model performance based on metrics such as accuracy. In this section, we discuss how these data uncertainty sources are pertinent to the discussion of uncertainty quantification along with literature resources for further investigation.

### Incomplete Data

Incomplete data refers to the phenomena where it is known (or suspected) that some potentially relevant data is missing from the dataset. This section highlights the extent in which the data may be incomplete: from missing values for features in a data sample, missing samples for part of the distribution, etc. Figure 8 provides a high-level outline of the manifestations of incomplete data.

### Missing data/values

It is not uncommon to encounter datasets that have values missing or unreported. The term **missing data** commonly refers to the phenomena where measurements of one or more features are missing from a sample. As part of the cleaning and preprocessing of data for the ML pipeline, a practitioner may easily identify the missing values and remediate their impact and/or account for this data uncertainty in their analysis. Missing data, in its various forms and respective treatments, is an active area of research in ML. The data may be missing at random (MAR), missing completely at random (MCAR), and missing not at random (MNAR). This classification of missing data has been adopted and used in causal reasoning literature [26][27].

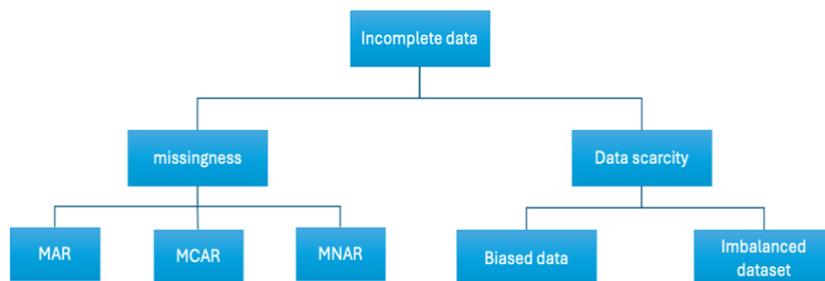

*Figure 8) Diagram showing the manifestations of incomplete data, as a source of data uncertainty. Missing values involves incompleteness at the scale of the samples (features missing within samples). Data scarcity refers to incompleteness of sections of the dataset (i.e. under representation of the population or imbalanced representation of classes).*

The reason behind missing values may vary, as some may be due to device fault, while others may be due to the nature of the data. Clinical/medical data is a good example of missing/incomplete data [26], as medical records are acquired at different points in time and may vary in the completeness of the measurements: for example while body temperature may be monitored every hour, blood pressure is monitored less frequently. This study points to an interesting missing value issue: medical records are acquired at a particular point in time, and there is high data uncertainty between these acquisitions of data. For example, a high spike in fever may be missed (not measured) if in between nurse visits.

Missing values is an undeniable source of data uncertainty; imputation methods and pre-processing techniques, designed precisely to address this issue, are active areas of research that have helped alleviate a lot of concerns with missing data. Imputation techniques (i.e. single value imputation like zero, mean, or median substitution) use existing samples to speculate the missing values. This uncertainty source falls under 'known unknowns', and we classify it as an epistemic uncertainty (can be reduced with more information) [29][30].

### Data scarcity

**Data scarcity** can be considered another form of missing data, in which areas of the data distribution are lacking samples (i.e. high epistemic uncertainty areas of Figure 5). Ultimately, the generalization capability of a model can be significantly influenced by the lack of data.

**Biased** data (i.e. underrepresented minority groups) and **imbalanced datasets** can also be considered forms of data scarcity and (by extension) missing data; biased data and imbalanced datasets refer to under-representation in the features of the samples and class distribution, respectively. In the case of bias, if a minority group is under-represented in the sample data, this will create a distribution that will not represent the overall population. [31] discusses potential ML algorithm biases when using electronic health record data; they explain how algorithms based on bias data may amplify already existing health care disparities. The implications and influence of bias (as a form of uncertainty) is an interesting and active research question [32][33]. For imbalanced datasets, some of the classes are over/underrepresented in the data distribution. As a potential source of uncertainty in the data, the influence and implications of imbalanced data on the predictive uncertainty has been a topic of interest [34][35]. The development of algorithms and techniques to compensate and alleviate imbalanced datasets is another active area of research [36][37].

### Noise

Noisy Data is another area of concern for data uncertainty. The amount of **noise** in the data samples can contribute to the perceived quality of the data and its usability for a given problem. [38] studies the sensitivity of different algorithms to measurement-noise. This study, dating from 2011, compares the noise sensitivity of four ML algorithms, namely decision tree, naïve bayes, support vector machine, and logistic regression. The study found naive Bayes to be most resistant to noise; however, with data improvement methods (noise reduction) decision trees performed better. This study highlights the importance of considering different ML modeling frameworks based on the data uncertainty, while mainly focusing on measurement noise. In a similar study, 12 learning algorithms are subject to study over their noise sensitivity on 12 diverse data sets [39]. The algorithms used are Support Vector Machines (SVM) with radial and polynomial kernels, Gaussian Process (GP) with radial and polynomial kernels, Relevant Vector Machines (radial kernel), Random Forest, Gradient Boosting Machines, Bagged Regression Trees, Partial Least Squares, and k-Nearest Neighbors. They found that gradient boosting machines have low noise tolerance, though its performance was comparable to other algorithms in lower noise levels. They also note that for gradient boosting machines, the fraction parameter influenced the noise sensitivity. [39] highlights the importance of understanding the noise present in the data and its possible influence on the ML model's performance.

Noise in the data, such as measurement noise or stochastic nature of the data, is mostly associated with aleatoric uncertainty (it is irreducible).

### False Labels

The trustworthiness of labels is another concern of data uncertainty, be it from false labels, label uncertainty, or noisy labels. Some researchers explore the classification challenges of datasets that have falsely reported labels (also referred to as **mislabeled data**) [40]. On the other hand, some labels may have uncertainty associated with them (known as **label uncertainty**), an example being labels for medical images that are hard to diagnose (even for domain experts) [41][42]. We consider **noisy labels** another example of how uncertainty can manifest in the data [43]. A survey of noisy-label robustness for different algorithms and deep learning methodologies is provided in [6]. They define noisy labels as labels that may have been corrupted (thus no longer reporting the ground truth), further explaining that real-world datasets have been reported to have a ratio of corrupted labels ranging from 8.0% to 38.5% [6].

### Data Scope

The **scope** of the (training and test) data and environmental uncertainty are another source of uncertainty that is its own niche field of study, especially in ML. This data uncertainty source is of a different nature in that it involves the choice of data scope (in contrast to investigating data quality). It requires a better understanding of the **environmental uncertainty**, basically checking if there are foreseeable changes in the conditions that should be considered (i.e. satellite images may be clear, and then contain obstructions by clouds) [6].

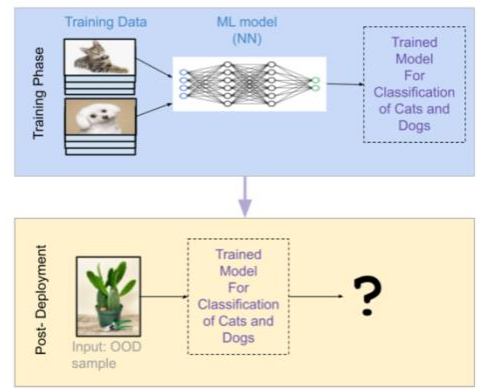

*Figure 9) The data scope problem: A model (blue box) is trained on a particular distribution (in this case, images cats and dogs). However, post-deployment, it is given a sample that is OOD (yellow box). The output for this OOD sample should, ideally, indicate that the model is uncertain since it has not seen such a sample before and can't make an accurate generalization.*

Part of the discussion of scope, **data shift** refers to when the trained model is given a sample that is not from the training distribution; the extent may range from in-domain, domain-shift, or out-of-domain (also referred to as OOD).

[6] limits discussions of data uncertainty to solely in-domain uncertainty. Thus, they consider the uncertainties in the data from the training set distribution. We would like to note that the quality and quantification of data uncertainty is influenced by the choice to include domain-shifted samples (i.e. blurred or mixed samples) or add more of them (i.e. data augmentation techniques). Hence, we decided to include scope and data-shift as sources of *data* uncertainty.

|  | In-domain | Domain-shift | Out-of-distribution |
| --- | --- | --- | --- |
| Input sample | Assumed to be equal to training distribution | Shifted version of the training distribution | Far and unknown from training data distribution |
| cat/dog classification example | Picture of a dog | Blurred picture of dog | Picture of a bird |
| Reducible with strategies | Yes | Somewhat | No |
| Desired behavior of UQ techniques | Low predictive uncertainty for correct predictions | Higher predictive uncertainty even for correct predictions | High predictive uncertainty; should abstain from prediction |

*Table 1) Data scope key insights adapted from [6]*

Domain-shift and OOD samples are a common usage and example of the added insight and value of uncertainty quantification. Ideally, the predictive uncertainty becomes worse (model is more uncertain) when dealing with domain-shift (also referred to as covariant shift) and OOD samples. This would allow human analysis and intervention for such samples, perhaps leading to changes in the model (i.e. reevaluation of the training distribution, etc.). Ultimately, to quantify uncertainty and communicate the value to ML practitioners, a metric is necessary. Table 1 provides examples for the predictive uncertainty, as well as the ideal behavior of the uncertainty estimations. For our example of an MNIST classification model, in the case that the input is an extremely low-resolution image of a number, in which the model can't decipher the key features of a number, we would want our output to provide a high predictive uncertainty, *regardless of whether it was able to accurately predict the number*. This is a key difference between an uncertainty-sensitive application vs the traditional focus on accuracy. In a standard ML practice, the focus may be more on accuracy (the model's ability to ultimately predict the correct class); however, in an uncertainty focus, the practitioner would like to know if the model is certain of its accurate prediction. For OOD samples, the ground truth 'label' is not available to the model; therefore, the model is not able to accurately make a prediction. For example, if the input to an MNIST trained model is a cactus, the model doesn't have a 'cactus' label since it hasn't seen this before. In such a setting, we desire the predictive uncertainty should reflect a very high level of uncertainty (or abstain from prediction).

For key insights into domain-shift and scope discussions, see Table 1. [1] and [6] are surveys of uncertainty quantification in DL with a focus on exploring the sensitivity of uncertainty estimation algorithms to shifts in domain. Interestingly, [1] found that traditional post-hoc calibration falls short under dataset shift, though not surprising (since the calibration data is also in-domain with the training data and can't capture the out-of-domain knowledge of a dataset shift). Along with their focus on distribution uncertainty, [6] looks at how the different applications of robotics, medicine and earth observation make use of uncertainty estimation techniques.

## Additional Notes and Discussion on Data Uncertainty

The information provided in this section gave the reader a look into possible sources and categories of uncertainty in the data, namely uncertainty sources that impact data quality and distribution. By human standards, a blurry image is considered low quality; it is considered a noisy sample, and thus high in data uncertainty. It is noteworthy that the relationship between data uncertainty (specifically sources that impact data quality) and the model performance (in terms of both prediction accuracy and uncertainty) is not always intuitive or straightforward. An example of this is data augmentation, which has been known to help some models, especially DL models, achieve better generalization results. Data augmentation (i.e. noise, MixUp) is a technique that intentionally includes variations of data, some of which have high data uncertainty. This may be surprising and unintuitive, as we are effectively adding low-quality data (in comparison to clear and crisp images of the classes). We refer the reader to the literature for more information on the influence of data augmentation on uncertainty estimates [30][31][44][45][46].

On a similar note, we should consider that some of the data uncertainty sources mentioned in this article are impacted by possible issues in the **management of data** and **data corruption**. The use of different media, un/structured data, various data sources, and different data collection techniques have led to the investigation of how emerging data handling requirements may influence the uncertainty in the data; for example, [47] is a study focused primarily on uncertain data management. Moreover, data corruption is possible through faulty data measurement, collection, or reporting [6]. Adversarial examples can also be considered a form of data corruption; [48] examines how uncertainty measures (i.e. specifically predictive entropy and mutual information) assist with detecting adversarial examples. Using probabilistic model ensembles, they demonstrate that uncertainty measures can help identify ambiguous data (high predictive entropy) and adversarial examples (mutual information).

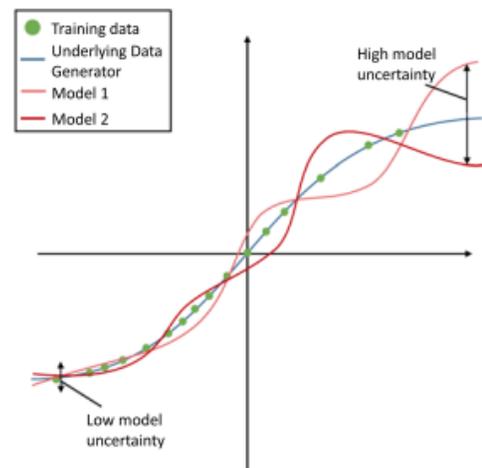

*Figure 10) (borrowed from [6]) Example of model uncertainty. Both Model 1 and Model 2 (pink and red lines, respectively) are trained on the same training data. Areas in which the models are similar, identify low model uncertainty. However, areas where the models differ significantly highlight areas of high model uncertainty.*

## 4.2. Model Uncertainty

We use the term model uncertainty to capture the uncertainties caused by the ML model. These uncertainties may stem from the choice of model framework (i.e. choice of logistic regression, deep neural networks, random forest, etc.), training procedure, hyper/parameter choices, as well as choice of inference strategies (if applicable to the framework, such as BNNs). [6] considers the model uncertainty to be the epistemic part of the predictive uncertainty; if the model had more information, it could give a more accurate representation of the data, so the model uncertainty is reducible. This is in juxtaposition with data uncertainty which [6] considers the aleatoric uncertainty segment of the predictive uncertainty (as it cannot be further reduced).

Figure 10 shows depictions of model uncertainty: areas of high model uncertainty show where the model needs more information to better represent the underlying data distribution, while areas of low model uncertainty indicate where the model is able to better learn the data distribution.

We recognize that there is literature focused on the uncertainty of the modeling process itself, such as research on modeling uncertainty estimations. Since this is not directly in reference to ML, we refer the reader to the literature [49].

### Framework Uncertainty

We consider framework uncertainty to capture the uncertainty associated with choice of ML framework (such as Logistic regression, Random Forest, Deep Learning, etc.) for the model. To the best of our knowledge, the literature for quantifying model uncertainty does not cover framework uncertainty. This means that estimations of model uncertainty do not account for possible model misspecification, for example. This is not surprising as concerns like framework uncertainty are difficult to capture, let alone quantify; [7] explains the in-feasibility of the meta-analysis (the statistical process of combining data from multiple studies to find common trends and results) needed to properly address and quantify framework uncertainty. Thus, further highlighting the –very valid– concern that this translates to disregarding some model uncertainty sources, such as model misspecification, in model uncertainty handling/quantification [7].

### Training Decisions

Training decisions refer to the decisions regarding training the model and preparing the data, such as pre-processing and cleaning the data, data augmentation, feature normalization, etc. An example of training decisions that is specifically relevant to our discussion are decisions to limit or modify the feature space of the data. In deterministic NNs, the complexity of the feature space, or lack thereof influences the model's ability to quantifying and distinguishing between the aleatoric and epistemic predictive uncertainty [9]. We saw this in Figure 6 which illustrates that the change in dimensions/feature space of the data can influence the (epistemic and aleatoric) data uncertainty. Although these decisions impact the (epistemic and aleatoric) data uncertainty, since the ML practitioner may decide to limit or modify the feature space of the data; this is part of the ML design and training decisions that can influence the uncertainty of the predictions. Thus, we consider it among the model uncertainty sources.

### Parameter uncertainty

Parameter uncertainty is defined as the uncertainty about the true parameter values in a model, and thus is enveloped in model uncertainty. Parameter uncertainty stems from the choices in hyperparameter and model design.

The weight distributions of the layers in a BNN are a good example of modeling parameter uncertainty. In BNNs, rather than use a single value for the weights, a distribution is used. After training the BNN model, the learned weight distributions can be a good indication of the parameter uncertainties for the weights of the model.

### Additional Notes and Discussions on Model Uncertainty

Understanding how some of the terms and concepts in the uncertainty literature relate to each other is one of the contributions of this paper. Hence, we would like to clarify the overlap in terms between model uncertainty and approximation uncertainty. Model uncertainty closely ties to what is referred to as 'approximation uncertainty' in the literature. Approximation uncertainty refers to the uncertainty of the model's approximation of the data distribution (or function) [7]. Model uncertainty, on the other hand, includes a broader range of uncertainties that are reflected in the prediction.

It should be noted that some of the literature considers the lack of samples or insufficient coverage of the training data as examples of model uncertainty, whereas we consider them as data uncertainty sources [50]. We acknowledge that the insufficiency of data has an impact on the model uncertainty (i.e. its ability to predict the data distribution); however, we consider this a propagation of data uncertainty sources into the model uncertainty.

## 5. Metrics: Uncertainty Quantification Metrics

The use of appropriate metrics is imperative to any study, especially in risk sensitive settings. As ML continues to grow, performance metrics have also evolved beyond error estimates and accuracy to better fit the requirements and concerns (i.e. accuracy is not a sufficient measure of performance with an imbalance dataset). [51] provides an overview of common AI evaluation metrics for medical applications, reviewing the merits of the different metrics and their added insights/pitfalls.

In the context of uncertainty quantification, however, we recognize that there are various forms of communicating and quantifying uncertainty, such as percentages, qualitative language (i.e. communicating areas of high/low epistemic uncertainty), etc. The literature reviewed for this study mainly uses probabilistic language (i.e. predictive distributions) for predictive uncertainty quantification, and alternative methods of uncertainty communication/quantification are not prevalent in the literature.

To the best of our knowledge, there is no universally accepted metric for uncertainty quantification; however, there are some metrics that have been more widely adopted in the literature. They are described briefly below. It is noteworthy that the evaluation of the quality of uncertainty estimates is challenging since quality depends on the underlying method of uncertainty quantification (UQ), ground truth for uncertainty estimates, etc. [6][52].

## 5.1 Metrics for quantifying predictive uncertainty

Based on the UQ approach and task (regression or classification), the metric for predictive uncertainty differs. In the case of UQ techniques that provide a predictive distribution (e.g., Bayesian Neural Networks), for a classification task, the predictive uncertainty is measured for a sample (metric: predictive distribution; entropy), or the overall predictive uncertainty of the trained model on the dataset (metric: entropy). For a regression task, the predictive uncertainty is quantified in a sample (metric: prediction interval) and overall uncertainty on the dataset (metric: standard deviation) [1][53][54][55].

For Deep Ensemble UQ techniques, the variance of the output of the ensemble is an uncertainty estimate [19]. For Dirichlet Prior Networks, [19] provides a comparison of different uncertainty measures (maximum probability, mutual information, expected pairwise KL divergence, and differential entropy) and provides a calibrated method of uncertainty estimation.

## 5.2 Metrics to evaluate the accuracy of the predictive uncertainty

Another concern in the discussion of uncertainty quantification and assessment is finding a measure to properly score the accuracy of the predictive uncertainty. This is a necessary metric in order to compare the UQ capabilities of different techniques and methodologies; and it is important to use a proper scoring rule (such as Brier score) [55]. To compare models under dataset shift, [1] uses the following metrics: predictive uncertainty, classification accuracy, negative log likelihood (on a held-out set), Brier score, expected calibration error, and confidence and predictive entropy for completely OOD inputs [1][53][54][55]. In another study, a proposed uncertainty accuracy metric incorporates the model prediction, ground truth label, and uncertainty value [54]. Their metric evaluates the overall accuracy of the uncertainty estimation as a ratio of the desired cases of the Negative Predictive value (conditional probability of correct prediction with low uncertainty) and True Positive rate (high uncertainty for incorrect predictions) over all possible cases [54]. In the proposed UQ technique of Epistemic Neural Networks (ENN), [56] suggests the use of KL-divergence with respect to the target distribution as a metric to compare different UQ techniques. Their testbed metric compares dropout, ensemble, a hyper model and Bayes by Backprop (BBB) as ENNs. According to this metric and their testbed evaluations, dropout shows poor performance, with bayes by backprop, hyper model and ensemble following [56]. This study further evaluates the performance (using the KL-divergence metric) of these UQ techniques with changes in data ratio, noise, and input dimensions. With respect to noise, ensemble performs well with low-noise, but BBB and hyper model perform well in environments with high-noise [56].

# 6. Assessment of Uncertainty: UQ methods for Deep Learning models

The variety of uncertainty sources, fluid uncertainty categories, etc. all contribute to the complexity of the task of communicating and quantifying uncertainty in DL (and ML in general). In this section, we provide a review of the literature that aims to assess and quantify predictive uncertainty in DL. This family of algorithms and techniques are referred to as uncertainty quantification techniques (UQ). While the literature provides substantial contributions to the field of UQ, to the best of our knowledge, no particular approach to uncertainty estimation has been declared the 'gold standard' and it is an ongoing research endeavor.

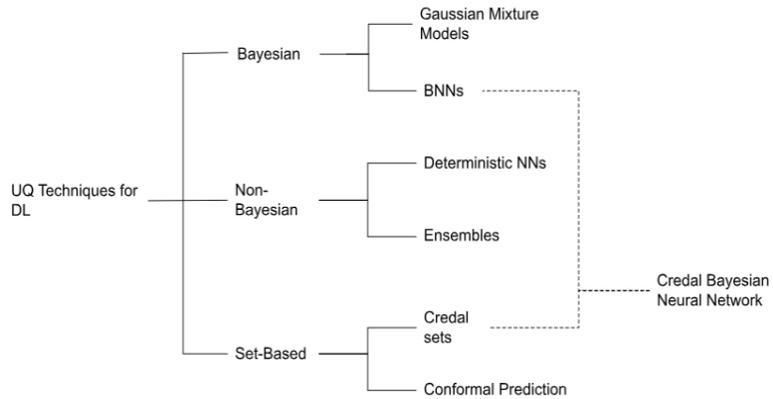

*Figure 11) Uncertainty Quantification techniques for deep learning models*

As the research in UQ evolves, we see more interactions between fields of study such as statistical schools of thought (frequentist vs Bayesian), imprecise probabilities (credal sets and Credal Bayesian Deep Learning), and ad-hoc uncertainty estimation techniques. Table 2 provides references to relevant UQ literature, for more information on their theoretical framework and applications. This section further includes a brief overview of the different UQ frameworks.

| Category of UQ Technique | | | Notable Works |
|---|---|---|---|
| Bayesian Methods | Gaussian Mixture Models | | Dual-supervised mixture of Gaussian Mixture Models [57], Gaussian Mixture Model for uncertainty propagation [58] |
| | BNN | Standard BNNs | Uncertainty correction (EpiCC) [59] |
| | | Variational Inference | (Stochastic Variational Inference) Initial adaptation [60], Bayes by Backprop [61], Reparameterization Trick [62] |
| | | | Normalizing flows [63][64] |
| | | | (Monte Carlo Dropout) MC-Dropout [65], MC-DropConnect [54], Heteroscedastic classification NN [14] |

| | | Sampling Methods | (Markov Chain Monte Carlo) Hamiltonian Monte Carlo [66], Stochastic Gradient Monte Carlo (SG-MCMC) [67], RECAST (SG-MCMC method) [68] |
|---|---|---|---|
| | | Laplace Approx. | Scalable Laplace Approximations [69] |
| | | Other BNNs | Bayesian Convolutional neural networks (CNNs) [70][91], Bayesian Recurrent neural networks (RNNs) [71] |
| Non-Bayesian Methods | Deterministic NNs | Internal | (Dirichlet) Prior Networks [16][72][73], Evidential NNs [74][75] |
| | | | Inductive Bias [9], Temperature-scaling Calibration [76], Inhibited Softmax [17], Radial Basis Function Networks (DUQ) [77] |
| | | | (Regression) Quantile Loss Function [78], Virtual Residual pre-training [79], Interval NNs [80] |
| | | External | Gradient Metrics [81][82][83], Additional NN for Uncertainty [84][42], Spectral-Normalized Gaussian Process (SNGP) [85] |
| | | Hybrid | Softmax & Feature Space Regularization (DDU) [86] |
| | Ensembles | Deep Ensembles | Automated deep ensemble with uncertainty quantification (AutoDEUQ) [87], Combination of Base and Meta model [55], (BNN Ensembles) Bayesian Ensembling [88][89], Approximately Bayesian Ensembling [90] |
| Set-based | Conformal Prediction | | Deep Conformal Prediction [92], Inductive Conformal Prediction [93], Conformal in NN [94], Conformal Prediction in CNN [95] |
| | Credal Sets | | Credal UQ [96], Credal Semi-Supervised Learning [97], Credal Bayesian Neural Network [98] |
| Other Literature | | | Test-time augmentation methods [99][100], Epistemic Neural Networks [56] |

*Table 2) Uncertainty Quantification Techniques and Relevant Literature*

## 6.1 Bayesian Methods and Bayesian Neural Networks

Bayesian approaches have led to various uncertainty estimation methods that have been the subject of a great deal of literature. The rapid usage and research in BNNs and Bayesian DNNs are quite obvious from a google search or simply asking ChatGPT about uncertainty quantification in ML. This is not surprising as NNs are also vastly used in modern applications, it follows that the Bayesian approach to UQ of NNs would be of increasing research interest.

The Bayesian framework is based on the Bayes' rule, which describes the probability of an event while considering prior knowledge that is related to the event. Using Bayesian inference, the posterior distribution is derived via Equation (1)

$$p(\theta|D) = \frac{p(D|\theta)p(\theta)}{p(D)} \quad (1)$$

Where $p(\theta)$ is the prior distribution, and the term $p(D|\theta)$ represents the likelihood of the data ($D$) being in the distribution predicted by the model with parameters $\theta$ [6]. The posterior distribution is therefore influenced by the prior choice and likelihood functions. Loss functions such as cross-entropy or mean squared error are examples that aim to achieve the maximum log-likelihood.

While the application of Bayesian approaches for UQ is not limited to BNNs (Gaussian Processes, etc.), our discussion is limited to the techniques utilized with DL in the literature:
- Gaussian Mixture Models
- Bayesian Neural Networks and Bayesian Deep Neural Networks
- Credal sets & Bayesian: Credal Bayesian Deep Learning

## Gaussian Mixture Models

[57] proposes a new framework, dual-supervised uncertainty-inference (DS-UI), and explains the capabilities of Gaussian layers for uncertainty inference. This framework combines the last fully connected layer of the classifier with Gaussian Mixture Models, providing a probabilistic interpreter for the input features. Their results show improvements in misclassification detection, as well as OOD detection. [58] studies the uncertainty propagation in neural networks with an adaptive Gaussian mixture method. They provide an adaptive Gaussian mixture scheme for highly accurate and faithful representation of the uncertainty, and computationally efficient estimation of uncertainty propagation.

## Bayesian Neural Networks and Bayesian Deep Neural Networks

Bayesian Neural Networks (BNNs) are NNs that incorporate Bayesian theory. BNNs take a prior and likelihood function, and incorporated within a neural network framework, the BNN learns an estimate of the weight distribution, rather than a single valued weight estimate in standard NNs. As a result, multiple passes for an input sample provides different prediction results. Together, these prediction results form a prediction distribution, which is desirable for UQ. BNNs are especially appealing as they combine the impressive capabilities of NNs (such as predictive performance, scalability, expressiveness) with Bayesian learning principles (i.e. prior belief) [6].

The fact that the posterior distribution as formulated in Equation (1) is an intractable problem, leads to the use of inference methods to find an approximation. There has been a significant amount of literature around the different inference methods and their success in providing good uncertainty estimations for the BNN. As mentioned before, the choice of inference method is a hyperparameter that the ML practitioner must choose. For more details see literature referenced in Table 2.

It is noteworthy, however, that BNNs come with drawbacks, such as high memory and computational cost, additional hyperparameter choices (i.e. inference approach, prior, likelihood function), and potential loss in accuracy. For example, when proposing Credal Bayesian Deep

Learning, the authors cite the potential shortcomings of regular BNNs choice of a single prior and likelihood, which may result in loss in robustness and miscalibration [101]. There has been fairly extensive research on the choice of prior and likelihood function, especially in the context of the cold posterior problem (where the BNN isn't able to learn the posterior distribution). Moreover, the posterior distribution of Standard BNNs doesn't allow for the capturing of epistemic uncertainty for each sample [7]. In more recent literature focusing on uncertainty calibrations, BNNs have been identified as providing overconfident and mis-calibrated uncertainty estimations for OOD samples [98].

[87] and [102] provide a survey of BNNs in different applications and the advantages of Bayesian deep learning. We refer the reader to [6] for an impressive listing of use cases, applications, and literature for Bayesian methods beyond uncertainty estimation, such as Bayesian model selection, model compression, active learning, continual learning, etc. Bayesian practices have also shown promising results in convolutional neural networks (CNNs) [70][91], Recurrent neural networks (RNNs) [71].

## 6.2 Non-Bayesian Methods

There are a number of UQ techniques that provide uncertainty estimates without utilizing Bayesian methods. The non-Bayesian methods provided below can be categorized as deterministic NNs and ensembles.

Deterministic NNs

Deterministic NN are neural networks that have deterministic parameters, thus repeated forward passes for a sample will give the same result [6]. Not surprisingly, UQ using deterministic NNs is an area of increasing research activity because of their relatively less computation and memory complexity (compared to BNNs, for example). However, Deterministic NNs rely on a single opinion/prediction (in contrast to the multiple predictions obtained in an ensemble, for example) [6]. As a result, Deterministic NNs are more sensitive to decisions in network architecture, training, and hyperparameters, which is a drawback of this method [6].

UQ using Deterministic NN can be categorized as internal methods and external methods (and a combination of the two). Internal methods require changes to the Deterministic NN architecture to learn the parameters of the prediction distribution (rather than standard Deterministic NNs which learn parameters for a single-point prediction). External methods, on the other hand, employ other techniques outside of the Deterministic NN architecture to complement the single-point prediction with uncertainty estimation. We have also added a hybrid category; these techniques have both internal and external components to achieve UQ with Deterministic NNs. A comprehensive list is provided in Table 2.

Deep Ensembles

Deep ensembles are very popular in uncertainty estimation literature. While their computation and memory complexity are prohibitive, they are still seen as one of the main methods to which most new algorithms are compared. [87], while making the case for Bayesian DL methodologies, clarifies that deep ensembles can be seen as approximate Bayesian marginalization, rather than a competing approach to Bayesian method. [55] proposes a deep ensemble that is created as follows: (1) select a proper scoring rule for training criterion, (2) smooth the predictive distributions using adversarial training, (3) create an ensemble of NNs. Deep ensembles, as demonstrated in [55], yield well-calibrated uncertainty estimates (comparable to approximate BNNs), as well as OOD detection.

AutoDEUQ, an automated deep ensemble with uncertainty quantification, seems like a natural progression in utilizing the UQ capabilities of deep ensembles [87].

## 6.3 Set-Based

Set-Based techniques characterize predictive uncertainty using sets (in contrast to single-point estimates). This approach can be deemed a response to the issue that predictive uncertainty may not always be adequately represented by a single distribution (or single prediction interval), but rather we need to account for multiple plausible outcomes [98]. Since we focus primarily on DL, we limit our discussion of set-based techniques to conformal prediction and credal sets.

Credal Sets

Credal sets represent sets of probability distributions and come from imprecise probability theory [7]. Instead of providing a single point estimate, machine learning algorithms can produce predictions in the form of credal sets, which are sets of probability distributions. Thus, credal sets provide a range of possible outcomes. Credal sets are often used when there is limited or incomplete information, which makes it an interesting candidate for uncertainty quantification in ML. Recent research has focused on the connection between machine learning and credal sets, providing an overview of uncertainty measures and their empirical comparison [96]. [98] proposes Credal Bayesian Deep Learning (CBDL), which utilizes credal sets of likelihood functions and priors. [96], [97], and [98] provide more information for credal sets in machine learning.

CBDL is an example of the use of credal sets for UQ in ML; this framework combines BNNs, credal sets, and ensemble strategies [98].

Combining Credal Sets & Bayesian: Credal Bayesian Deep Learning

Credal Bayesian Deep Learning (CBDL) is an interesting combination of BNNs with imprecise probabilities (namely credal sets) [98]. Rather than select a single prior and likelihood function, CBDLs use a credal set of priors and likelihood functions. As such, the BNN is trained with all combinations of the set of priors and likelihood functions. The predictive probability distribution (posteriors of the different BNNs) provided by these credal sets are then combined into a convex hull, which is ultimately the credal posterior set. These posteriors, and the convex hull itself, illustrate the ambiguity of choosing the correct prior and likelihood distribution. The theoretical background and details of the capabilities of CBDLs are provided in [98]. CBDLs prove to be more robust to distribution shifts than standard BNNs, and they provide a means to distinguish and quantify epistemic and aleatoric uncertainties [98]. [98] explains that the predictions from CBDLs 'most probably contain the correct prediction'. CBDLs are compared to Ensemble BNNs, with the experiments showing CBDLs to outperform the ensemble [98].

### Conformal prediction

Conformal Prediction is considered a frequentist framework that gives an uncertainty estimate for the predictions using a calibrator [29] [30]. When using conformal prediction, a calibration set is set aside prior to training. This calibration set is used to observe how well the trained model predicts the samples (compares predictions to ground truth values). These observations enable the calibrator to measure how well the model will predict similar samples in the future/test set; this translates to prediction intervals. The prediction intervals give an indication of the amount of uncertainty associated with a prediction (i.e. a larger interval means the higher uncertainty for the prediction for input). These prediction intervals are adjustable based on a predefined confidence level. Essentially, the probability of excluding the correct target sample is bounded by this confidence level [7].

[92] is a study of Deep conformal prediction for Robust Models, with the proposed method providing good performance with noisy data. For information on variations of conformal prediction, namely Transductive Conformal Prediction, Inductive Conformal Prediction, Density-based Conformal Prediction see [92]. We refer the reader to the literature for more details on Conformal prediction and its variations [103].

## 6.5 Other Uncertainty Estimation Methodologies

For a more complete inclusion of UQ methodologies, we have included a few additional methods here. Most of these approaches are not used in a lot of the literature but have been included in reviews. They are viewed to have significant drawbacks that we view as their main reason for not being adopted in recent literature and comparisons. For the sake of completeness, they are included here, along with the literature in which the reader may find more details.

In 2018 survey, [104] focuses on uncertainty quantification methodologies in Neural Networks, primarily on methodologies that provide prediction intervals. According to their review, there are significant drawbacks to the Delta Method and Mean-Variance Estimation Method (MVEM) making them less suitable for most applications (such as computational complexity, and underlying assumptions such as uniform noise and predictive power of the NN). As such, we do not include them further in the review of UQ methodologies, beyond listing them here.

Test-time augmentation methods, on the other hand, utilize a single deterministic network, but use data augmentation techniques on the input data at test-time. By doing so, several predictions are generated that are used for uncertainty evaluation of the prediction. As mentioned in [6], this method is relatively simple as the underlying model and data doesn't change, and data augmentation is achieved with available libraries. [6] explains the need to ensure that the augmented data is meaningful and within the target distribution. The study provided by [100] is interesting in that it finds that test-time augmentation can change the accuracy of the model (change many correct predictions to incorrect predictions, and vice versa); they propose a learning-based test-time augmentation method to overcome this issue. [6] highlights that one of the significant open research questions around test-time augmentation is the influence of different kinds of augmentation techniques on uncertainty.

Epistemic Neural Networks (ENN) is a framework in which existing UQ techniques can be framed, where the quality of the ENN is evaluated using KL-divergence [56]. It utilizes an epistemic index. Where BNN research aims to find probabilistic inference tools for NNs, ENNs attempts to identify NNs that are suitable tools for probabilistic inference.

Some other uncertainty estimation techniques make use of other fields of research, such as physics-informed uncertainty quantification [105].

## 7. Model and Uncertainty Calibration

One of the discussions around ML models is whether they are well calibrated. Model calibration is intertwined with uncertainty quantification, as it is deemed necessary for proper uncertainty quantification [6]. Calibration is also a concern of uncertainty literature, aiming at uncertainty calibration. As part of a complete discussion of uncertainty, this section outlines the relevance between model calibration and uncertainty, as well as uncertainty calibration.

### 7.1 Model Calibration

Model calibration is explained as the process(es) that is applied to an ML model to improve its probability estimates [106]. A well-calibrated model will have predicted probabilities that closely match the true probabilities (i.e. ground truth correctness likelihood); one of the common measurements for model calibration is expected calibration error (ECE) [107]. Model calibration is also sometimes referred to as confidence calibration [107]. For an interesting investigation on the different processes for model calibration and miscalibration observations (such as the influence of regularization, capacity, etc. on model calibration), we refer the reader to [107].

There are a number of techniques for NN calibration that involve different levels of the ML process: data, model, or post-processing. Namely, [6] categorizes the NN calibration techniques as regularization methods (training phase), NN uncertainty estimation methods (model and framework decision phase), and post-processing methods (after training). Interestingly, this study surveys these NN calibration techniques as uncertainty calibration techniques [6]. We refer the reader to [6] for a survey of these methods. It is noteworthy that BNNs are considered an uncertainty calibration method in [6]; however, overconfident uncertainty estimations is considered one of the shortcomings of BNNs [20].

## 7.2 Uncertainty Calibration

The literature uses the term uncertainty calibration in various ways. Some studies refer to model calibration efforts and measurements as uncertainty calibration [108] [20]. For example, [108] provides a deeper study of the necessity of calibration of uncertainty estimates in UQ techniques, highlighting that adaptivity and consistency are complementary concepts for validating uncertainty estimates. On the other hand, some researchers discuss uncertainty calibration as calibration efforts on uncertainty estimates [109].

In another study, the term 'uncertainty calibration' is used to refer to calibration under uncertainty, thus requiring a differentiation on calibration assessment in high-uncertainty settings [19]. [19] defines uncertainty calibration as: a well-calibrated model would have an increase in ECE and uncertainty measure correlating with added uncertainty (here, mutilated data). Hence, for their evaluation of calibration uncertainty, both the ECE and uncertainty measure(s) are investigated as the mutilation of the data increases [19]. It is noteworthy that [110] refers to this as Calibration Under Uncertainty (CUU). They define CUU as the adjustment of parameters to reflect data and model uncertainties.

Uncertainty calibration, as one may intuitively expect, can also refers to the calibration of uncertainty estimates. This usage of the term 'uncertainty calibration' better aligns with the goal of understanding and dealing with uncertainty. Ultimately, when using uncertainty estimates for risk-critical decisions, the calibration of these measurements and their alignment with the ground truth is of utmost importance. [48] explains the necessity of assessing the calibration of uncertainty estimates from NN uncertainty quantification techniques. Their calibration technique utilizes the following uncertainty measures: aleatoric and epistemic uncertainty, mutual information, and entropy.

## 8. Decision Making and Communication of Uncertainty

It is important to understand how uncertainty is communicated and used in the decision-making process. The form and content of uncertainty should be compatible with the nature of the decision it supports, as well as the requirements and constraints defined by the decision maker. For example, a qualitative expression of uncertainty may be deemed inadequate by a decision maker. This is particularly important in the context of high-impact decisions (i.e. the safety and/or risk concerns). UQ and how they relate to decisions are an interesting area of research. An example of this would be a focus on how the uncertainty estimations may influence decisions/prescriptions of drug dosage, lead or determine alternate course-of-action or options (i.e. Reinforcement Learning), etc. [71] also notes that there are decision making theories, such as three-way decisions and info-gap decisions, which can be useful when dealing with uncertainty. However, in our review, we do not see these decision-making theories employed.

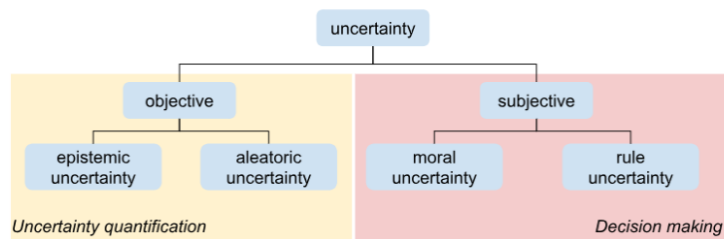

*Figure 12) Encyclopedia taxonomy for uncertainty. (adapted from [111])*

For a deeper understanding of the role of decision making under uncertainty, [111] explains the relationship between risk and uncertainty. While [111] is an encyclopedia entry and doesn't involve the discussion of ML directly, the general taxonomy it provides is applicable (Figure 12). While the UQ focuses on the objective sections of the taxonomy, the subjective section reflects the decision-making uncertainties. Moral and rule uncertainty are considered subjective uncertainties in this taxonomy, meaning that they are not quantifiable. Moral uncertainty, which is an intuition guided decision, reflects the moral implications of decisions made or informed by AI systems and ML [111]. Rule uncertainty refers to rule-guided decisions, which is also subject to uncertainty [111]. [112] explains that rule-guided decisions are involved when there is a lack of a moral rule; hence, there is a fallback on rules to govern and make decisions. Both of these 'subjective' uncertainties play an important role in risk sensitive applications, making the understanding and decision making based on uncertainty an important discussion.

## 9. Conclusion and Discussion

We established that uncertainty handling and estimation methods are a rapidly developing field in the introductory section of this paper. While providing an updated survey of the literature surrounding UQ techniques for DL, we discussed the uncertainty sources, categories, metrics and final decision-making process. We believe this paper, and its comprehensive approach to addressing the myriad of topics around uncertainty, represents a necessary step towards organizing and synthesizing various discussions surrounding uncertainty.

We complete our paper with a compilation of open challenges identified in the literature, thus providing grounds for future work [5][1][6]:

**No cohesive taxonomy, baseline, and framework for UQ:** For ML practitioners that desire to include uncertainty estimation techniques into their models, there is no clear and comprehensive taxonomy to understand the uncertainties in their specific ML process, as well as how to best handle the uncertainties. This shortcoming, admittingly, goes hand in hand with the lack of a systematic comparison framework, insufficient implementation of baselines, and lack of metrics for comparing different uncertainty estimation methodologies. In 2021, Google took an initial step in addressing this gap with its open-source baseline library [113].

**No Uncertainty ground truths:** It remains difficult to validate existing methods due to the lack of uncertain ground truths. There is no standard dataset with uncertainty ground truths to evaluate the accuracy of predictive uncertainties. To the best of our knowledge, [54] is among the few works that address the lack of uncertainty ground truth in their uncertainty metric proposals. However, further research is needed to find an approximation for the uncertainty ground truth. [6]

**Mapping to sources of uncertainty:** Ultimately, it would be useful for practitioners to be able to locate the sources of uncertainty in their data, model, and overall ML process, allowing them to map the impact of some of these uncertainties in their relative sources and overall model performance. Moreover, the handling and quantification of some sources of model uncertainty (i.e. framework uncertainty) remains a research challenge.

**Epistemic and Aleatoric uncertainty quantification:** Several studies and software packages focus on calculating some of the aleatoric and epistemic uncertainties in the prediction. The underlying principles of some of these studies, such as the additive properties of measures of epistemic and aleatoric uncertainty for the total uncertainty, is called into question in recent literature [50]. [50] explain that the use of conditional entropy and mutual information is an increasingly common way of quantifying aleatoric and epistemic uncertainty, respectively, in ML. However, their findings raise concerns about the adaptation of these measures in the current state-of-the-art. In their discussion of emerging UQ techniques, [7] further emphasizes the desire for a mathematically rigorous decomposition of the total measure of uncertainty into aleatoric and epistemic parts. Thus, it remains an interesting research opportunity. We consider the breakdown of uncertainty quantification of the aleatoric and epistemic uncertainties, as they relate to the sources of uncertainty, another interesting avenue for future work.

**Explainability of uncertainty estimation models:** Uncertainty estimations are an important step towards explainable AI; explainable uncertainty estimations would give an even deeper understanding of the neural network's decision process. This enables the practical deployment of DNNs to incorporate risk-averse capabilities. [114] is interesting research in approaching explainability assisted with uncertainty; they show close ties to estimations of aleatoric and epistemic uncertainty. Figure 13 illustrates how uncertainty quantification techniques can assist explainable AI, through the identification of areas of AI confusion/ambiguity.

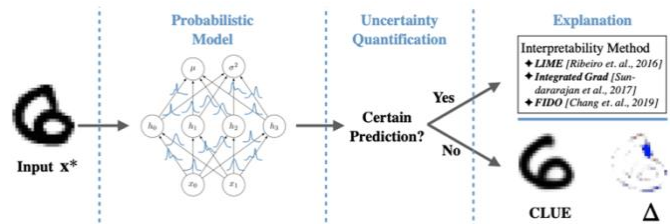

*Figure 13) (borrowed from [114]) CLUE framework identifying the features that create class ambiguity*

**Small datasets:** Most uncertainty quantification methods are suitable for large datasets. It remains an interesting field of research to find uncertainty handling techniques that are better suited for small datasets. This is especially useful since labeling datasets, especially in fields like medicine and health care, are expensive and access to large datasets is limited.

**Time, Numeric & Computational optimization:** Finding less computationally complex, memory expensive, and/or scalable ways to use UQ is an active area of future research. Ultimately, reducing the computational and memory costs, while retaining the same performance, for UQ is a key research challenge. [6][1]

**Other possible research avenues for UQ:** A list of areas in which there has been limited investigation of uncertainty handling/quantification include semi-supervised and self-supervised learning models, UQ for other non NN related ML frameworks (i.e. Random Forests), fusion-based methods, alternative decision-making theories, efficient Bayesian Optimization, efficient UQ for graph neural networks, continual learning, among others.